\documentclass[letterpaper, 10 pt, conference]{ieeeconf} 
\IEEEoverridecommandlockouts                              
\usepackage{amsmath} 
\usepackage{algorithmicx}
\usepackage{algpseudocode}
\usepackage{graphicx} 
\usepackage{amsmath}
\usepackage{color}
\usepackage{subcaption}
\usepackage{verbatim} 
\usepackage{url}
\usepackage{cite}

\newcommand{\unit}[1]{\ensuremath{\mathrm{#1}}}

\title{\LARGE \bf
Finding the Easy Way Through
-- \\the Probabilistic Gap Planner for Social Robot Navigation 
}

\author{Malte Probst$^{1}$, Raphael Wenzel$^{1}$, Tim Puphal$^{1,2}$, Monica Dasi$^{1}$,
\\Nico A. Steinhardt$^{1}$, Sango Matsuzaki$^{3}$, Misa Komuro$^{3}$
\thanks{$^{1}$Honda Research Institute EU GmbH {\tt\tiny first.last@honda-ri.de}}%
\thanks{$^{2}$Honda Research Institute JP Co., Ltd. {\tt\tiny tim.puphal@jp.honda-ri.com}}%
\thanks{$^{3}$Honda R\&D Co., Ltd. {\tt\tiny first\_last@jp.honda}}%
}
\begin{document}

\maketitle
\begin{abstract}
In Social Robot Navigation, autonomous agents need to resolve many sequential interactions with other agents. State-of-the art planners can efficiently resolve the next, \textit{imminent} interaction cooperatively and do not focus on longer planning horizons. This makes it hard to maneuver scenarios where the agent needs to select a good strategy to find gaps or channels in the crowd. We propose to decompose trajectory planning into two separate steps: \textit{Conflict avoidance} for finding good, macroscopic trajectories, and \textit{cooperative collision avoidance} (CCA) for resolving the next interaction optimally. We propose the Probabilistic Gap Planner (PGP) as a conflict avoidance planner. PGP modifies an established probabilistic collision risk model to include a general assumption of cooperativity. PGP biases the short-term CCA planner to head towards gaps in the crowd. In extensive simulations with crowds of varying density, we show that using PGP in addition to state-of-the-art CCA planners improves the agents' performance: On average, agents keep more space to others, create less tension, and cause fewer collisions. This typically comes at the expense of slightly longer paths. PGP runs in real-time on WaPOCHI mobile robot by Honda R\&D.
\end{abstract}

\section{Introduction}

\noindent The goal of Social Robot Navigation (SRN) is to enable robots to seamlessly navigate public, unstructured, and potentially crowded environments. While this task is seemingly simple for humans, it poses significant challenges for autonomous agents \cite{mavrogiannis_core_2023}. Despite many important abstract questions like intention estimation, politeness, or contextual appropriateness \cite{francis_principles_2023}, avoiding collisions with people in a cooperative, foresighted manner without being overly cautious remains an active topic in SRN research.

State of the art SRN algorithms can provide short-term plans of a few seconds in real time, which enables them to navigate through dense crowds effectively \cite{macenski_desks_2023}. 
Despite these recent achievements, planners with shorter horizons have difficulties to create foresighted behavior. 
Cooperative collision avoidance (CCA) for longer horizons remains difficult, due to higher uncertainties and, when coupling planning and prediction, the resulting combinatorial explosion of potential future outcomes in more crowded environments \cite{mavrogiannis_core_2023, sun_mixed_2024, singamaneni_human-aware_2021, khambhaita2020viewing}.

To address this problem from a slightly different angle, we follow a simple observation by \cite{mirsky_conflict_2024} in the context of \textit{conflict avoidance} for social navigation. The authors define a conflict between two agents as ``a situation in which, if there is no change of direction or a change in speed by at least one of the parties, they will collide``. Furthermore, the authors note that ``not all conflicts end in a physical collision, but every collision is preceded by a conflict``. Accordingly, in this paper, we propose to split the task of motion planning on a trajectory level for SRN into two distinct planning stages: First, a long-term planner which reduces the number of conflicts by finding foresighted, macroscopic trajectories, such as deviating slightly but early to evade an oncoming pedestrian, or finding potentially suitable gaps in a crowd of people. Second, a short-term planner which avoids collisions by resolving microscopic cooperative interactions.
We make two main assumptions in our approach: First, we assume that the long-term planner can rely on the short-term planner to handle (cooperative) collision avoidance. Second, we assume that all other agents (e.g., pedestrians) also want to avoid collisions - at least to some extent. Given this premise, the long-term planner can anticipate some degree of cooperativity from other agents it will encounter in the future.
Together, the assumptions relax the trajectory planning problem for the long-term planner significantly, since it does not have to discard conflicting trajectories at all costs, or model each potential future interaction in a detailed manner (e.g., by explicitly coupling prediction and planning).
\begin{figure}
    \centering
    \vspace{0.2cm}
    \includegraphics[trim={1.5cm 0 0.5cm 1.5cm},clip,scale=0.66]{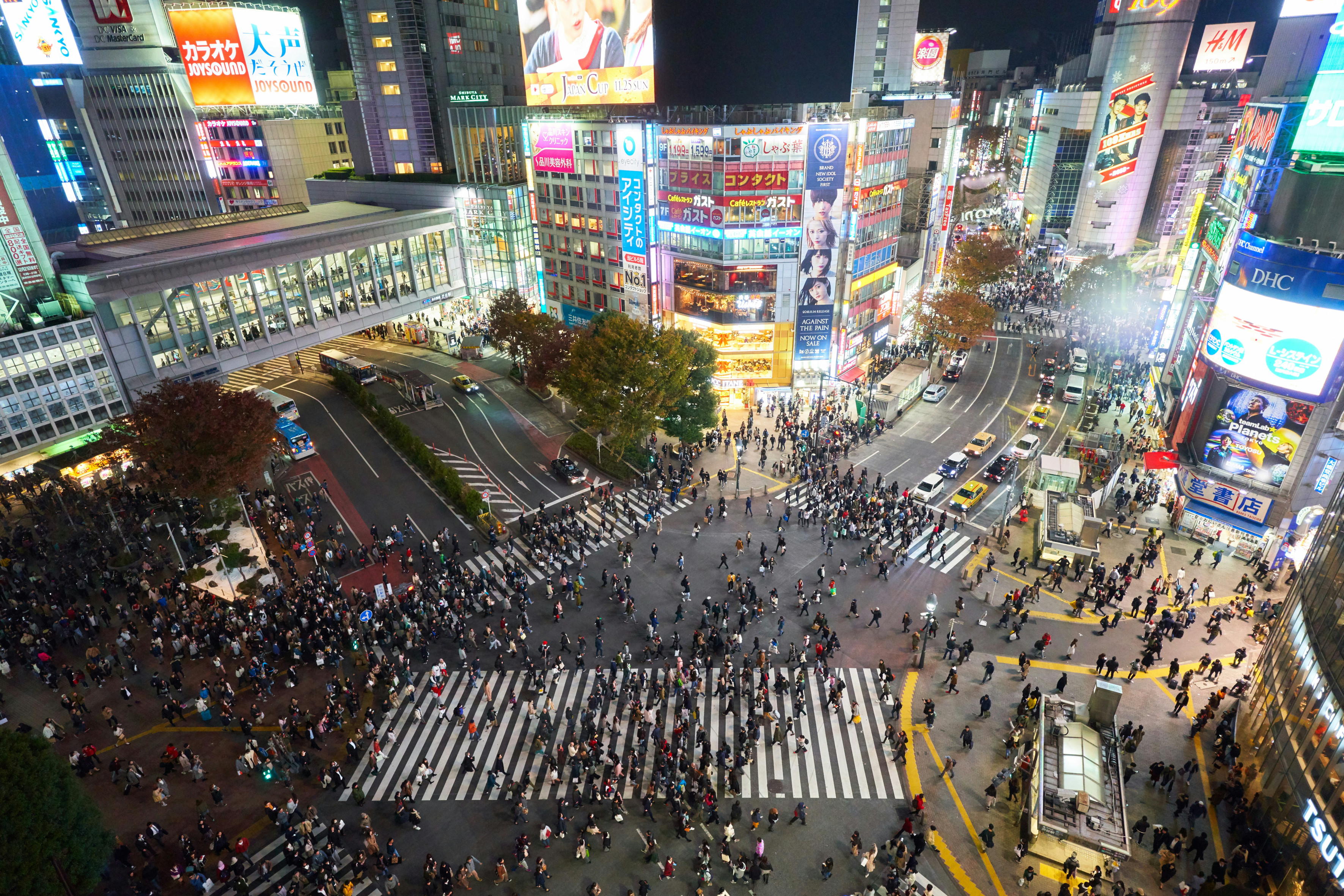}
	\caption{Autonomous robots in public places like Shibuya Crossing, Tokyo need to be able to navigate smartly in crowded, unstructured environments. Here, areas with varying density can emerge which the robot can leverage to improve its motion through the crowd. \textit{Photo by T. Volz}}
	\label{srn}
\end{figure}
This paper presents the Probabilistic Gap Planner (PGP) as a long-term planner. PGP applies a well-known probabilistic risk model proven in both the AD and the robotics context \cite{eggert_continuous_2017}.
We modify the probabilistic risk calculation to include PGP's cooperativity assumption. PGP guides the short-term planner towards areas with lower predicted density, i.e., less crowded trajectory options on the way towards the goal. PGP relies on the short-term planner to resolve all microscopic interactions without collisions.

In extensive experiments in simulation, we assess how adding PGP on top of state-of-the-art short-term planners improves the performance in scenarios with varying levels of crowdedness. The results show that PGP helps to find significantly less dense paths through crowds of people and reach the goal faster without compromising safety. Furthermore, we test PGP in indoor corridor scenarios on the WaPOCHI mobile robot by Honda R\&D, where it runs in real time.

\section{Related Work}
\noindent Hierarchical planning is a common approach in autonomous driving (AD). Often, the problem is decomposed into route planning, path planning, maneuver planning (or decision making), and control \cite{varaiya_smart_1993,gonzalez_review_2016,claussmann_review_2020}. The maneuver planning and control layers operate on trajectory level, and typically consider dynamic objects. For example, common approaches use lattice planners or discrete search techniques for coarse maneuver planning, and numerical optimization for control considering the vehicle dynamics. Due to the structure imposed by roads, lanes, or traffic rules it is often feasible to define high-level maneuvers or even parametrized maneuver primitives such as "lane change" or "right turn".

In robotics, hierarchical planning is also quite common. However, many architectures on real robots use a combination of path planning and collision avoidance \cite{macenski_desks_2023}. It is less common to have multiple layers of trajectory planning, especially in the scope of the maneuver space. One reason could be that the planning problem is inherently less structured, since robots can, in principle, use the full 2D space for navigation.

Nevertheless, some approaches use multiple trajectory planners. \cite{kant_planning_1988} combine an s-t graph-based collision avoidance planner with a with local planner superimposing corrections based on sensed obstacles. 
\cite{bergman_improved_2021} combine a coarse lattice planner with numerical optimization for trajectory fine tuning, mainly focusing on static objects. Both approaches consider trajectories with predicted collisions infeasible and remove them already in their respective hierarchical layer.

In the context of SRN, a recent research focus has been on cooperatively solving interaction problems. Cooperative Collision Avoidance (CCA) algorithms go beyond standard reactive collision avoidance (CA) algorithms like DWA \cite{fox_dynamic_1997} by including assumptions about cooperativity. For example, ORCA \cite{siciliano_reciprocal_2011} enables real-time, collision-free navigation for multiple agents by computing velocity constraints that ensure reciprocal, cooperative collision avoidance while optimizing individual motion objectives. \cite{khambhaita2020viewing} include cooperativity aspects in a planning algorithm based on braids, a representation used to describe trajectories of multiple agents in a shared space over time. \cite{le_social_2023} augment a model predictive control algorithm with a sophisticated learned prediction model, thereby coupling prediction and planning. \cite{sun_mixed_2024} use iterative Bayesian updates over probability distributions to solve a mixed-strategy Nash equilibrium, finding cooperative solutions in real time.

However, many of the algorithms which explicitly consider trajectory-based interactions between agents face the problem of exponential complexity when more than a few other agents need to be considered, due to larger planning horizons or higher crowd density \cite{khambhaita2020viewing, singamaneni_survey_2024}. By design, they typically focus on the next, \textit{imminent} interaction at any point in time.

Summarizing, while there are approaches that use multiple hierarchical trajectory planners, they typically remove potentially colliding trajectories early on, which makes them susceptible to the \textit{freezing robot} problem \cite{trautman_unfreezing_2010}. In the context of SRN, there are a number of effective CCA algorithms. However, they typically consider the next interaction with a few other agents at a time, rendering larger planning horizons or denser crowds computationally challenging.

\section{Probabilistic Gap Planner}
\label{pgp}
\noindent PGP should avoid or reduce potential conflicts with other agents in the planned trajectory, assuming general cooperativity of agents. While PGP prefers planned trajectories with fewer conflicts, it does not perform collision avoidance, but relies on the CCA planner for this task. This relaxes planning and prediction significantly. Therefore, unlike most collision avoidance algorithms, PGP does not render a trajectory with a high level of predicted conflict, i.e., potential collision risk, infeasible. Instead, it reduces the expected utility of the trajectory, depending on the collision probabilities encountered along the way (see \ref{risk_theory}). Therefore, PGPs main goal is to maximize the expected utility of the planned trajectory with respect to the long-term goal. To achieve this, PGP creates a set of candidate trajectories (\ref{predictions}), evaluates the expected utility of each trajectory in the set (\ref{pgp_evaluation}), and, from the best one, extracts a subgoal for the CCA planner.%

\subsection{Cooperative Collision Risk and Survival Analysis} 
\label{risk_theory}
\noindent PGP employs the probabilistic collision risk model of \cite{eggert_predictive_2014, damerow_risk-aversive_2015,eggert_continuous_2017}. This approach has been used in the AD and ADAS context for collision avoidance (see, e.g., \cite{puphal_probabilistic_2019, probst_automated_2021}) and calculates the integrated probability that a planned ego trajectory will collide with any of multiple other predicted trajectories in the future.

The model assumes that the location $X_{i,k}$ of agent $k$ at future time $i$ is a normal distribution with the predicted mean $\mu_{i,k}$ and standard deviation $\sigma_{i,k}$: 
$$X_{i,k} \sim \mathcal{N}(\mu_{i,k},\,\sigma^{2}_{i,k})$$
Typically, $\sigma_{i,k}$ grows over the predicted time to reflect the growing uncertainty as to where an agent (ego or other) will be located in the future. Accordingly, the probability of two agents colliding, i.e., occupying the same space at time step $i$ is defined as the product of their two Gaussian distributions (see \cite{puphal_probabilistic_2019, probst_automated_2021}).

In practice, we calculate the probability $p_{i,k}^\text{coll}$ of the ego agent colliding with other agent $k$ at time step $i$ as
\begin{equation}
\label{collprob}
p_{i,k}^\text{coll} = 
\exp(-\frac{||\mu_{i,\text{ego}} - \mu_{i,k}||_2^2}{2\tilde{\sigma_i}}) * \frac{\tilde{\sigma_0},k}{\tilde{\sigma_i},k}
\end{equation}
with
$$
\tilde{\sigma_{i,k}} = \sigma^{2}_{i,\text{ego}}+\sigma^{2}_{i,k}
$$

Note that the exponential term of Eq. \ref{collprob} equals one if the agents' predicted location distributions' means are identical. The second term normalizes the probability to the initial variances $\sigma_{0,\text{ego}}$ and $\sigma_{0,k}$ such that $p_{0,k}^\text{coll}=1$ at $t=0$, and $p_{0,k}^\text{coll}<1$ for $t>0$ if the variances grow over time. In other words, the collision probability typically decreases with growing variances, i.e., decreasing probability density.

However, the risk model assumes that the random variables are \textit{conditionally independent}, whereas one of PGP's underlying assumptions is that agents are, in general, \textit{cooperative}. That is, they will tend to avoid future collisions by changing course, speed, or both.
In turn, this means that the predicted location distributions are, in fact, \textit{not} conditionally independent and that the actual probability that the agents will be colliding is typically lower.

Estimating the true collision probability of dependent location distributions for two agents is difficult and would require coupling planning and prediction. Therefore, we introduce an \textit{approximate} normalization factor $\tau\leq1$ as
\begin{equation}
\label{coop_normalization}
\tau = \frac{\sigma_{0,\text{ego}}}{\sigma_{i,\text{ego}}} * \frac{\sigma_{0,k}}{\sigma_{i,k}}
\end{equation}
In effect, $\tau==1$ if the predicted collision is imminent ($i==0$), or if both standard deviations remain constant over time (i.e., we are very certain about both agents' future locations). In all other cases, $\tau$ decreases linearly with growing standard deviation of either agent, since agents have more space to evade, i.e., cooperate.\footnote{Note that it is conceivable that we could parametrize the growth of $\sigma_{i,k}$ in a way that Eq. \ref{collprob} already possesses the desired properties, rendering $\tau$ unnecessary. Nevertheless, we argue that it is easier to \textit{explicitly} capture the general assumptions about an agent's motion in the growth of $\sigma_{i,k}$, and specifically include the approximated cooperation assumption via $\tau$.}

Finally, using $\tau$, we get the predicted \textit{cooperative} collision risk as
\begin{equation}
\label{coop_collprob}
p^\text{coll, coop}_{i,k} = p^\text{coll}_{i,k} * \tau
\end{equation}

Similar to the collision probabilities with dynamic objects, we calculate the probability $p^\text{coll, static}_i$ of colliding with a static object at time $i$. To keep things simple, we model the collision probability using the minimal lateral distance $d_i$ of any static object to the ego agent and use a 1d Gaussian for ego's location distribution. Consequently,
\begin{equation}
p^\text{coll, static}_i = 1-\text{erf}(\frac{d_i}{\sigma_i\sqrt{2}})
\end{equation}
with erf$(z)$ being the Gauss error function defined as 
$$
\text{erf}(z) = \frac{2}{\sqrt{\pi}}\int_{0}^{z}e^{-t^2}\text{d}t
$$

Having calculated the predicted static object collision probability and cooperative collision probabilities, we now use the Survival Analysis from \cite{eggert_continuous_2017}. It integrates $p^\text{coll,static}_i$ and all $p^\text{coll, coop}_{i,k}$ from multiple other agents $k$ over time to get the probability $p_i^\text{surv}$ of "surviving" up time step $i$ by converting all probabilities to event rates (for more details refer to \cite{eggert_continuous_2017}).
For PGP, we will use the survival probability $p_i^\text{surv}$ to discount the predicted utility of a planned trajectory in the evaluation (see Section \ref{pgp_evaluation}).

Considering the discrete case, the Survival Analysis calculates the total survival probability $p_i^\text{surv}$ of the ego agent as
\begin{equation}
\label{survival}
p_i^\text{surv} ~= \prod_{j<i} \exp(-\sum_{0}^{k}{p^\text{coll, coop}_{j,k}}+p^\text{coll, static}_i+p^\text{escape})
\end{equation}
Here, $0<p^\text{escape}<1.0$ is the escape probability, which can be interpreted as the probability of an unforeseen event happening at any time step. In practice, this term will make the $p_i^\text{surv}$ decay exponentially over time, and fall to zero in the limit, even in the absence of any static or dynamic collision risks.
Any non-zero collision probability with another agent or a static object will lead to a stronger decrease of the survival probability at time step $i$. As a consequence, if a trajectory has a high collision risk at time step $t$, e.g., by leading through a dense crowd of people or close to a static object, it's expected utility at $t=j\ge i$ will be discounted stronger, compared to a trajectory without any close encounters (see Section \ref{pgp_evaluation}).

\subsection{Predictions and PGP Trajectory Search Space}
\label{predictions}
\noindent PGP uses a classic predict-than-plan approach, decoupling prediction and planning. PGP's prediction step uses simple constant-velocity-constant-angle extrapolations of all other sensed agents.
\label{search_space}

\begin{figure}
    \centering
    \includegraphics[scale=0.55]{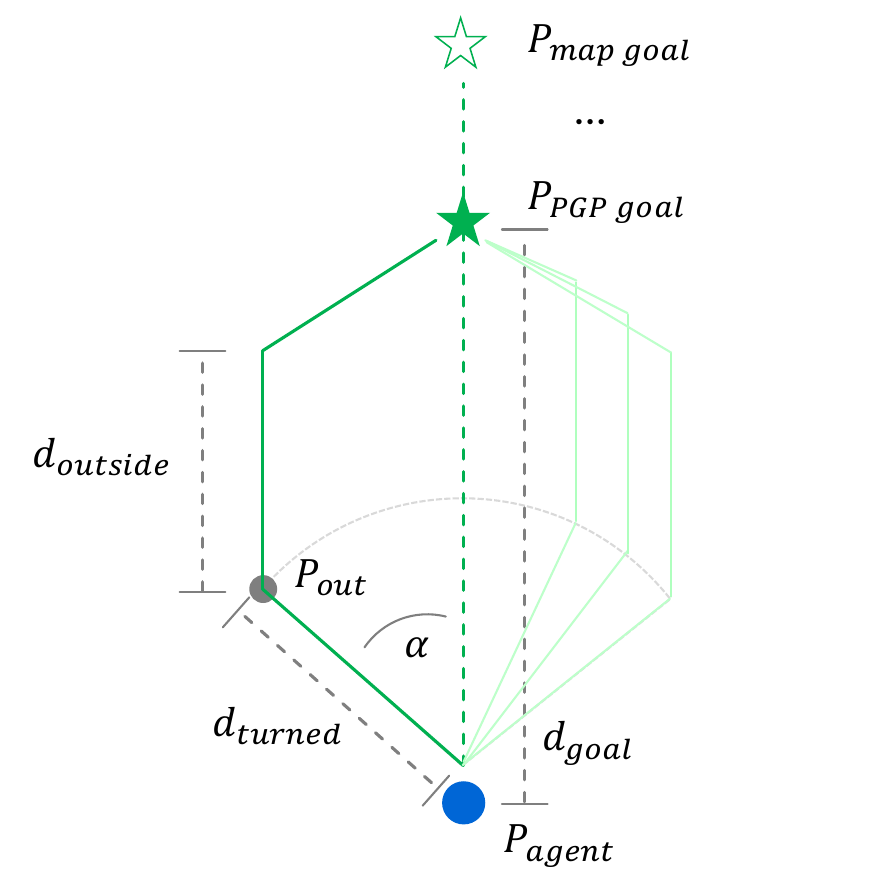}
    \caption{PGP path creation. The agent (green dot) wants to reach the map goal (light green star). PGP calculates path variations around the straight line to the goal to create candidate trajectories.}
\label{PGP_paths}
\end{figure}

The total space of potential plans is very large, containing all kinematically possible ego trajectories within the prediction horizon. Here, PGP uses a fixed sample of trajectories to cover potentially promising areas of the search space. The sample is based on variations of the direct path from the current ego position to the long-term goal (see Figure \ref{PGP_paths}). The quality of each trajectory with respect to the robot's goal will be evaluated later. First, we derive a feasible goal for PGP within its prediction horizon $T$, at a distance of $d_\text{goal}=T*v^\text{max}$, assuming $T$ seconds of constant maximum velocity $v^\text{max}$ towards the goal on the map $P_{\text{map goal}}$. Then, we create paths consisting of three segments. The first segment of length $d_\text{turned}$ is directed away from the straight path to the goal at angle $\alpha$. The second segment of length $d_\text{outside}$ is, again, parallel to the straight line to goal. The third segment connects the second segment to the goal.

For each path, we create a constant velocity trajectory with $v=1.0$ \unit{\frac{m}{s}}. We accommodate for the fact that the agent cannot turn arbitrarily fast by reducing velocity for a short time if considerable turning (i.e., more than 30$^{\circ}$) is required. In that case, we approximate the required time to turn as $t^\text{turn} =\frac{\gamma^{\text{turn}}}{\dot\gamma^\text{max}}$, with $\gamma^{turn}$ being the angle required to turn towards $P_\text{out}$, and $\dot\gamma^\text{max}$ being the maximal turn rate. Then, we reduce the initial velocity for $t^\text{turn}$ seconds to $0.5\cdot v$. Accordingly, the final set $L$ for evaluation contains trajectories fanning out at different angles in both directions, staying "outside" for a while or returning back immediately.

\subsection{PGP Trajectory evaluation, selection, and goal passing}
 \label{pgp_evaluation}
\noindent PGP evaluates the expected utility $u^{\text{expected}}_l$ of each trajectory $l$ in the set of all trajectories $L$ by summing up all expected utilities over the planned horizon.
For each time step $i$, we calculate the utility $u_{i,l}$ based on the planned velocity $v_{i,l}$ and the angle difference $\beta_{i,l}$ between the heading and a straight line towards the goal $P_\text{PGP goal}$ as:
\begin{equation}
u_{i,l} = \frac{v_{i,l}}{v_\text{max}}* \frac{\text{cos}(\beta_{i,l}) + 1}{2}
\end{equation}
The second term normalizes $u_{i,l}$ to be between zero (heading away from goal) and one (heading towards goal).

We now weigh the utility by the probability of reaching this time step in the first place (see Eq \ref{survival}).
Therefore,
\begin{equation}
u_{i,l}^\text{expected} = p^{\text{surv}}_{i,l} * u_{i,l}
\end{equation}
\begin{equation}
u_l^\text{expected} = \sum_{i}{u_{i,l}^\text{expected}}
\end{equation}

PGP now selects the trajectory $l^*$ with the highest expected utility $u_{l*}^\text{expected}$ , and uses it to calculate a subgoal which is passed to the collision avoidance planner. To calculate the subgoal, we simply use the trajectory's planned initial heading $\alpha_{l^*}$ (see Section \ref{search_space}), and place the subgoal at a distance $d_\text{subgoal}=v^\text{max, CCA}*t^\text{plan, CCA}$, with $v^\text{max, CCA}$ and $t^\text{plan, CCA}$ being the desired velocity and planning horizon of the CCA planner. Hence, PGP biases the CCA planner to go into the initial direction of $l^*$.

\section{Experimental setup}
\subsection{Scenario and Evaluation}
\begin{figure}
    \vspace{0.12cm}
    \centering
    \includegraphics[scale=0.47]{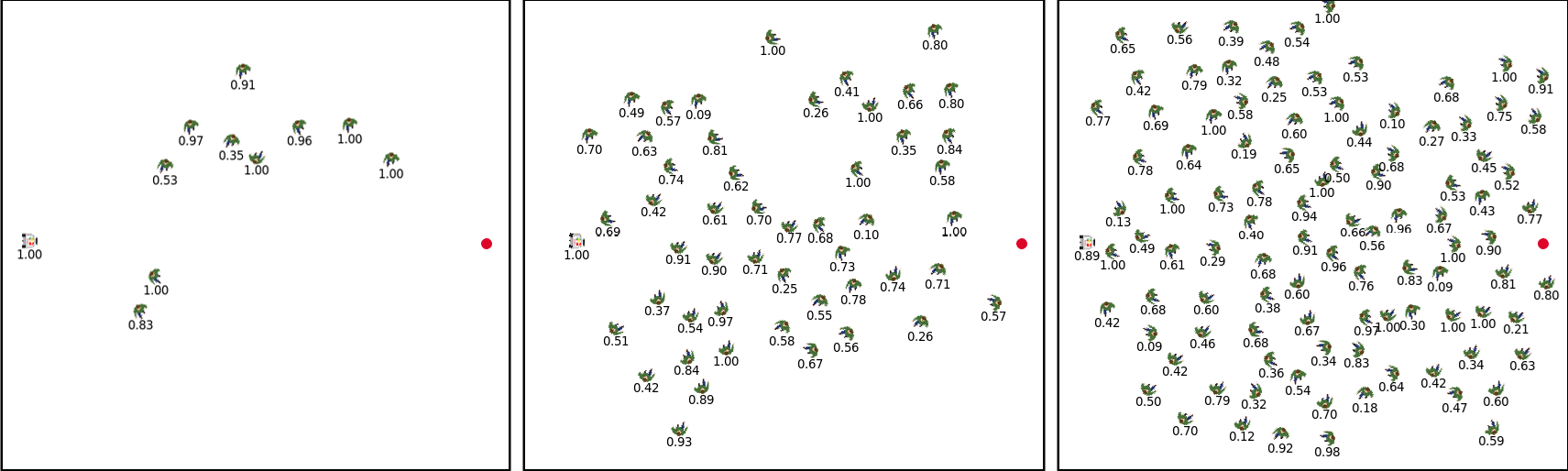}
    \caption{Screenshots from three simulated scenarios with varying density (from left to right: 0.1, 0.5, and 1.0 agents / $\text{m}^2$). The agent (robot icon) needs to traverse a 10 x 10 m stage towards a goal (red dot) populated by groups of other agents moving around to random goals on the stage.}
    \label{scenario}
\end{figure}
\noindent We evaluate PGP in an open space scenario with a varying number of pedestrians (see Figure \ref{scenario}). The scenario is conceptually similar to the \textit{Random} scenario of \cite{wang_metrics_2022} and the \textit{Crowd Navigation} scenario of \cite{francis_principles_2023}. Concretely, the agent needs to traverse a quadratic stage of 10 x 10 m. Other pedestrians are spawned with random start and goal locations within the stage. All agents have a desired velocity of 1.0 \unit{\frac{m}{s}}. We simulate all other agents in their groups with the Social Forces model \cite{helbing_social_1995}.\footnote{We use the Social Forces implementation from \url{https://github.com/yuxiang-gao/PySocialForce}} We vary the density of other agents from $0.01 \frac{\text{agents}}{\text{m}^2}$ (one single other agent) to a maximum density of $1.0\frac{\text{agents}}{\text{m}^2}$. Note that densities $> 0.72 \frac{\text{agents}}{\text{m}^2}$ already fall in a range where "the majority of persons would have their normal walking speeds restricted and reduced due to [...] avoiding conflicts" and the kind of scenario is "representative of the most crowded public areas, where it is necessary to continually alter walking stride and direction to maintain reasonable forward progress" \cite[p. 8]{fruin_designing_1970}. To impose more structure on the flow of pedestrians, we create groups of up to four people. Approximate spawning and goal locations are shared within a group.
When any agent of a group has arrived at its goal, we sample a new goal on the stage for the whole group to avoid standing agents blocking the stage. The preferred walking speed is $1.0\frac{\text{m}}{\text{s}}$ for all agents, including the ego agent.

With the variable density, the complexity of the scenario can vary drastically. It can be comparably simple, with only a few other agents that the ego agent needs to cooperate with. On the other extreme, it can be very complex, with many other agents, where continuous flow is interrupted, dynamic channels are forming and collapsing, and agents may get stuck due to suboptimal short-term decisions (or, simply, bad luck).

We evaluate the performance of the ego agent with the following metrics (see \cite{francis_principles_2023, gao_evaluation_2022}):

\subsubsection{Success Metrics}
We evaluate the performance by measuring the time the ego agent needs to reach the target, the total path length, and the collision rate while it is moving.\footnote{We assume that if an agent is standing, it should not be responsible for avoiding collisions.}
\subsubsection{Quality/social Metrics}
We calculate the Space Violation Rate (SVR), also referred to as Space Compliance in \cite{francis_principles_2023}. SVR is the rate at which the agent is closer to any other agent than a given threshold (here, use a threshold of 1.0m). Same as for the collision rate, we only calculate the SVR when the agent is moving.
Second, we use the Social Force model to calculate the average sum of Social Forces, i.e., the repulsive force between pairs of agents, affecting the ego agent. In contrast to threshold-based metrics such as SVR, the average Social Force also captures aspects of how a conflict evolves over time. We therefore interpret it as a proxy for social compliance, observing that the average force between two agents typically decreases faster if both agents act towards resolving the conflict, instead of only one agent performing all evasive actions alone.

\subsection{Tested algorithms}
\noindent As baseline agents, we use three common approaches. First, we use an agent implementing the Dynamic Window Approach (DWA) \cite{fox_dynamic_1997}. DWA is a pure CA algorithm, and does not make any assumptions about cooperations. We use a \textit{predictive} version of DWA, i.e., collision checks with others are based on their linear predictions. Second, we use an agent using Optimal Reciprocal Collision Avoidance (ORCA) \cite{siciliano_reciprocal_2011}. ORCA assumes that agents cooperate for collision avoidance by taking "half of the responsibility for avoiding pairwise collisions". As a third baseline, we use a Social Forces (SF) agent \cite{helbing_social_1995} which is frequently used for crowd simulations. The underlying idea is that agents' movements are subject to a number of pulling and pushing forces.

To test our PGP algorithm, we extend each of the baselines with an upfront PGP planner. Specifically, in each time step, we run one PGP planning step. We then use the output of PGP, i.e., a goal in x/y coordinates, and give this goal to the respective collision avoidance algorithm.

Therefore, we evaluate the performance of three baseline agents (DWA, ORCA, SF) and three PGP-enabled agents (PGP+DWA, PGP+ORCA, PGP+SF).

\subsection{Parametrization}
\subsubsection{Basic parameters}
For PGP, we use a prediction horizon of 8.0 seconds at 4 Hz, totaling in 32 planned samples. The agent's maximum velocity is 1 $\frac{\text{m}}{\text{s}}$.
\subsubsection{Risk calculation}
We parametrize the growth of all agents' Gaussian location distributions as follows: 
The time-dependent standard deviation $\sigma_i$ starts from an initial $\sigma_{0}=0.1666$, i.e., nearly all the probability mass is within an 1m radius around the agent's predicted position. $\sigma_i$ increases over the predicted time, with a linear dependency on the predicted velocity at any particular time step. Larger velocities lead to larger increases, assuming that fast agents tend to show larger deviations than slow or standing agents. The total standard deviation is capped at a speed-dependent value $\sigma_\text{max}$. Specifically,
$$\sigma_{\text{max}} = \text{min}(\beta*\sigma_\text{0}, \sigma_{0}+\gamma*\text{max}_i(v_i))$$
and
$$\sigma_{i} = \text{min}(\sigma_\text{max},\sigma_{i-1}+\delta*v_i)$$
In order to let $\sigma_i$ stay constant over time at standstill, and grow to three times the initial value over the total prediction horizon for the maximum velocity, we choose 
$\beta=3.0$, 
$\gamma=0.4$, and
$\delta=0.015$ for all agents.

\subsubsection{PGP Trajectory creation}
PGP creates the set $L$ of trajectories with the following parameters: A goal distance $d_\text{goal}=8.0\text{ m}$, fan-out angle $\alpha \in [-80, -64, \dots , 64, 80]\text{ deg}$, a segment length pointing outward of $d_\text{turned}=2.5\text{m}$. For each fan-out-angle $\alpha$, we create two paths, one which immediately turns towards PGP's goal, the other one stays "outside" for 90\% of the remaining distance before turning back, i.e., $d_\text{outside}\in [0,0.9*||P_\text{out}-P_\text{PGP goal}||]\text{m}$.

\subsubsection{Simulation}
We run simulations on stages with increasing population $\text{density} \in [0.01, 0.1, 0.2, \dots , 1.0] \frac{\text{agents}}{\text{m}^2} $, from a single other agent up to a total of 100 other agents, corresponding to one agent per $m^2$ on the 10 x 10 m stage. For each algorithm and density, we run 100 randomized scenarios, with identical random seeds across the evaluated algorithms.

\subsubsection{Benchmark algorithms}

The main parameters for the benchmark algorithms DWA, ORCA, and SF are listed in Table \ref{tab:parameters}.

\begin{table}[ht]
    \centering
    \begin{tabular}{|p{3.9cm}|p{2.0cm}|p{1.42cm}|}
        \hline
        \textbf{DWA} & \textbf{ORCA} & \textbf{SF} \\
        \hline
        $t^\text{plan}=2.0\text{s}$, $\text{dt}=0.25\text{s}$, \newline
        $v^\text{min}=0\frac{\text{m}}{\text{s}}$, $v^\text{max}=1\frac{\text{m}}{\text{s}}$,\newline
        $|\dot{\psi}|^\text{max}=1 \text{rad}$,
        $|\ddot{\psi}|^\text{max}=1.5 \frac{\text{rad}}{\text{s}^2}$,
        $|a|^\text{max}=1.5 \frac{\text{m}}{\text{s}^2}$,
        dynamic window grid: 10 yaw rates * 10 velocities
        &
        $t^\text{plan}=2.5\text{s}$,\newline
        $\text{dt}=0.25\text{s}$,\newline
        $v^\text{max}=1\frac{\text{m}}{\text{s}}$,\newline
        $d^\text{neighbor}=4.0$m,\newline
        $n^\text{neighbors}=5$,
        &%
        $v^\text{max}=1\frac{\text{m}}{\text{s}}$,\newline
        $\text{A}=5.1$,\newline
        $\lambda=3.0$,\newline
        $\gamma=0.35$,\newline
        $n=1$,\newline
        $ n^\prime =3.0$
        \\
        \hline
    \end{tabular}
    \caption{Main parameters of benchmark algorithms; $\dot{\psi}$: yaw rate; $t^\text{plan}$: planning horizon}
    \label{tab:parameters}
\end{table}

\subsection{Results}

\begin{figure}[h!]
    \centering
    \begin{subfigure}{0.49\textwidth}
        \centering
        \includegraphics[trim={0.5cm 0 0.5cm 0.5cm},clip,scale=0.51]{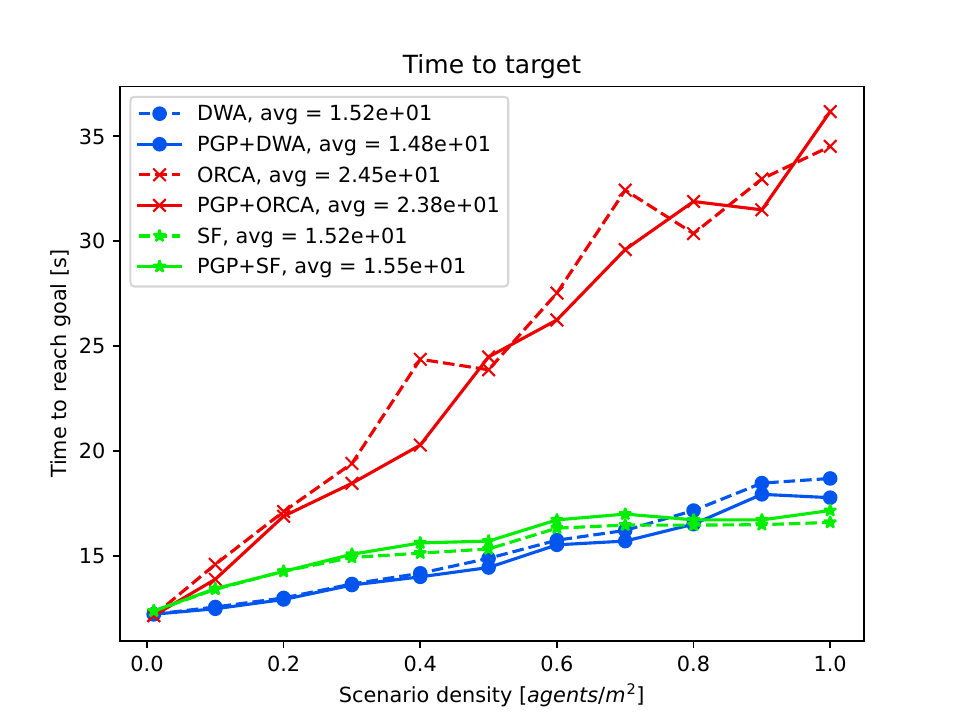}	
        \label{time_to_target}
    \end{subfigure}
    \begin{subfigure}{0.49\textwidth}
        \centering
        \includegraphics[trim={0.5cm 0 0.5cm 0},clip,scale=0.51]{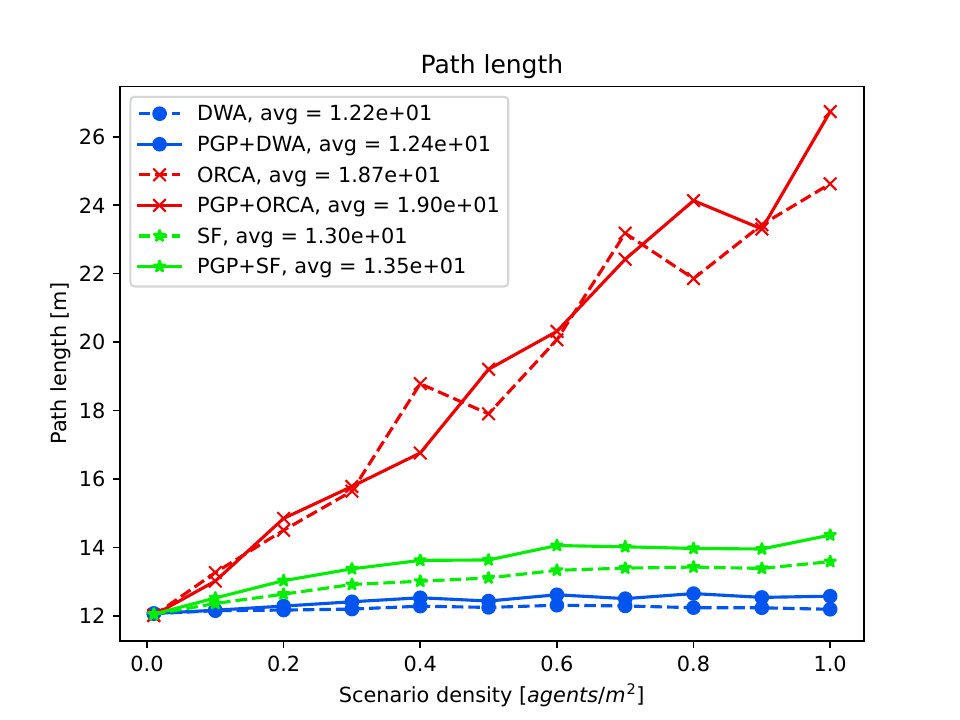}
        \label{path_length}
    \end{subfigure}
    \begin{subfigure}{0.49\textwidth}
        \centering
        \includegraphics[trim={0.3cm 0 0.5cm 0},clip,scale=0.51]{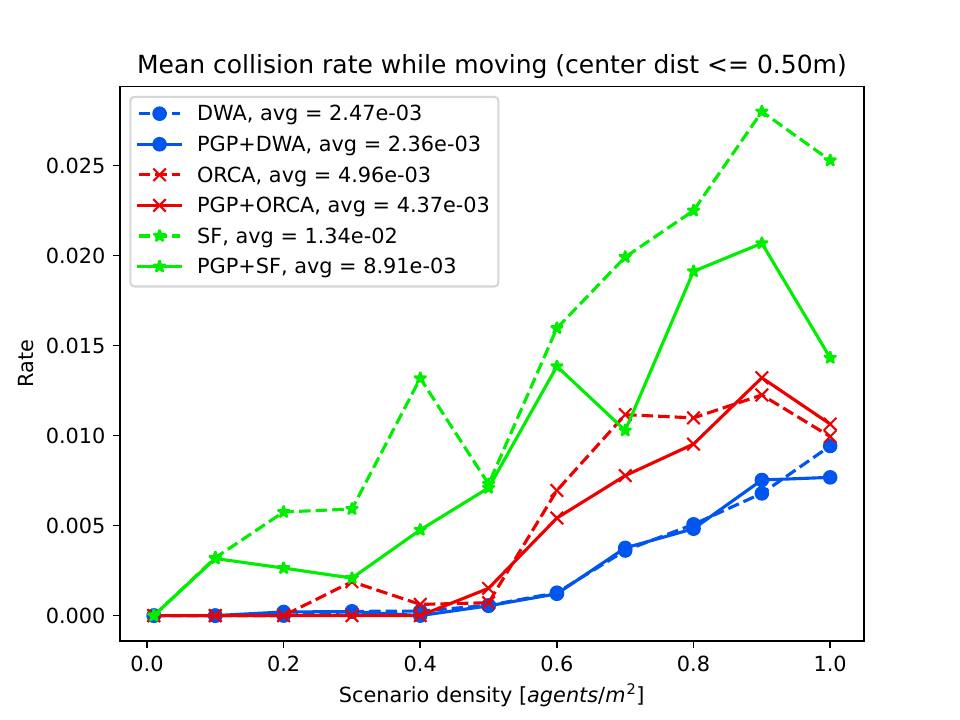}
        \label{collisions}
    \end{subfigure}
    \caption{Success metrics for open space scenario with increasing agent density.}
    \label{results}
\end{figure}

\begin{figure*}
    \centering
    \begin{subfigure}{0.49\textwidth}
        \centering
        \includegraphics[trim={0.5cm 0 0.5cm 0.2cm},clip,scale=0.51]{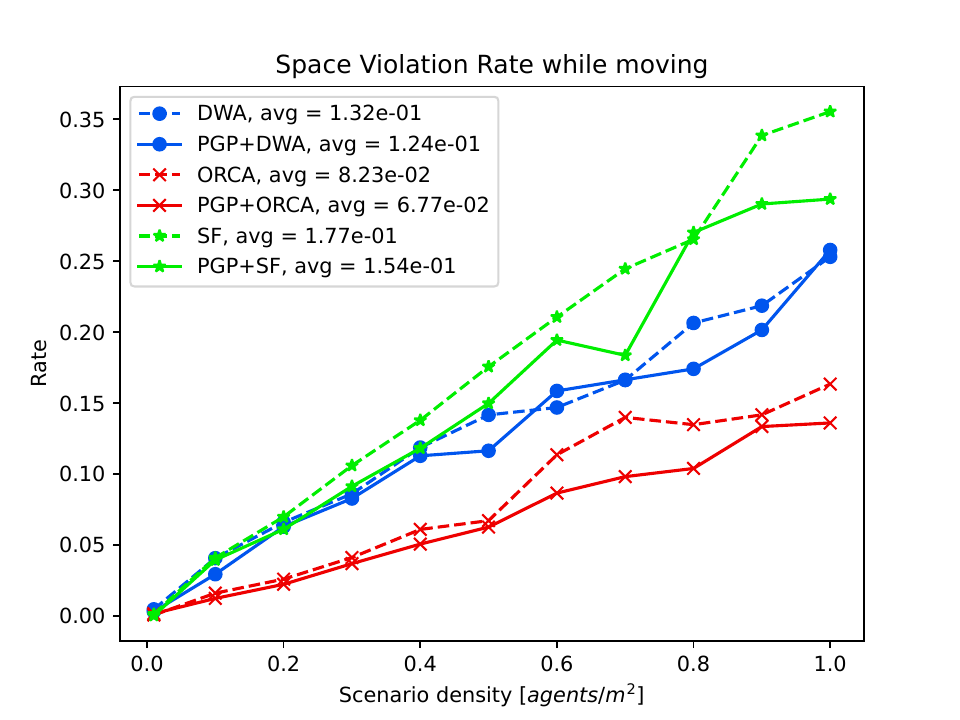}
        \label{space_violation}
    \end{subfigure}
    \begin{subfigure}{0.49\textwidth}
        \centering
        \includegraphics[trim={0.5cm 0 0.5cm 0.2cm},clip,scale=0.51]{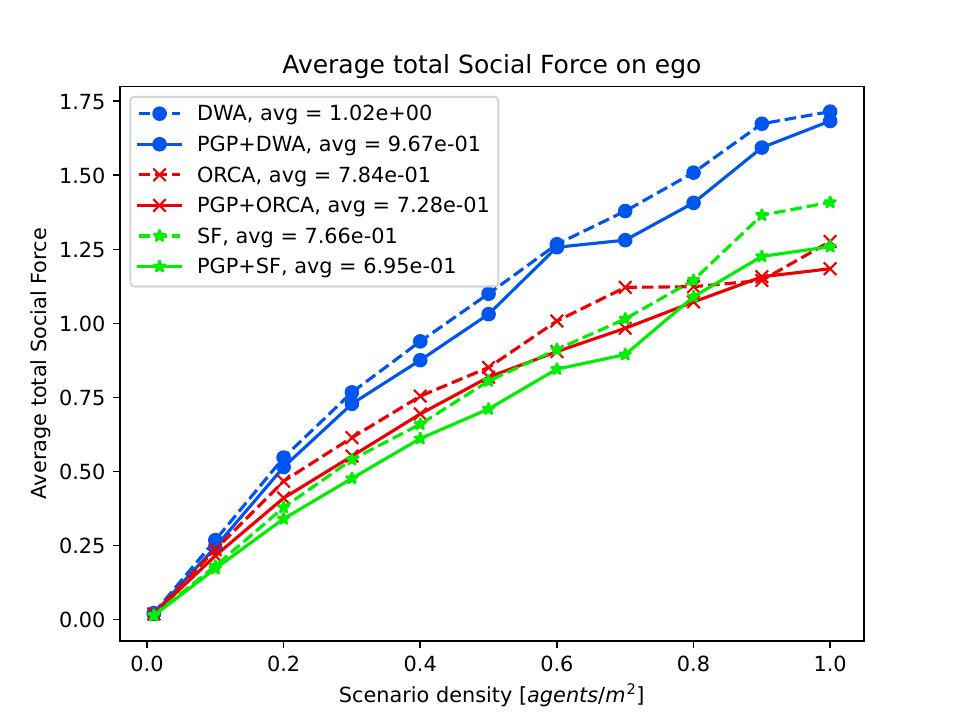}
        \label{forces}
    \end{subfigure}
\caption{Quality/social metrics for open space scenario with increasing agent density.}
\label{results2}
\end{figure*}
\noindent In the following, all differences are statistically significant, using a Wilcoxon-signed-rank tests with $p\le 0.05$.

The performance metrics in Figure \ref{results} show that all three (cooperative) collision avoidance algorithms (DWA, ORCA, SF) lead to very different behavior in the scenario. However, adding a PGP planner to the agent (solid lines) generally improves the behavior of the agent compared to using the standalone (C)CA planners (dashed lines).

The DWA algorithm (blue lines, solid and dashed) deviates very little from the straight path towards the goal (middle plot). It shows the smallest collision rate across the three (C)CA algorithms, with only a moderate increase for the highest density scenarios (bottom plot).

With the addition of PGP, DWA is able to significantly reduce the average time to reach the goal from approximately 15.52 to 14.48 seconds across densities, with larger advantages for more crowded scenarios (top plot of Figure \ref{results}). 
At the same time, the overall average collision risk (where DWA already has the lowest total levels) decreases further from 0.25\% to 0.24\%. 

The ORCA algorithm (red lines, solid and dashed) needs more time to reach the goal, compared to DWA and SF (top plot). This is especially true for denser crowds. In our simulations, ORCA agents tend to make large detours to avoid other agents, resulting in a significant increase in total path length with increasing density (middle plot). At the same time, ORCA shows a relatively low collision rate while moving (bottom plot).

For PGP+ORCA, the average change from 24.5 seconds to 23.8 seconds is not statistically significant. However, in terms of collision rate (bottom plot), PGP reduces the collision slightly for ORCA (0.44\% vs 0.50\%)

The SF agents (green lines, solid and dashed) also show a lower time-to-target than ORCA (top plot). For lower densities, SF agents are slower than DWA agents, but for the most crowded scenarios, they are slightly faster. Social Forces agents also stay mainly on-course, with a slightly longer path length compared to DWA (middle plot). Their fast pace, however, comes at the cost of a significantly increased collision rate compared to the other algorithms (bottom plot).

Combining the SF agents with PGP, the agent gets marginally slower (15.5s vs 15.2s).
However, in terms of collision rate (bottom plot), PGP drastically reduces the collision while moving for the SF agent (0.89 \% of time steps vs. 1.34 \%).

Note that the reduction of collision risk (lower plot of Figure \ref{results}) only results from the biased goals for the (C)CA algorithms. The middle plot shows that, as expected, PGP significantly increases the average path length of the agents. This is not surprising, since without PGP, the (C)CA planners target is always the true goal, whereas PGP biases them to deviate from the straight line.

Next, we look at the quality/social metrics of the agents (Figure \ref{results2}).
As expected, all three (C)CA algorithms get closer to other agents in the high-density scenarios (left-hand plot). The Space Violation Rate while moving is lowest for ORCA (recall that it often takes a long way around), highest for SF, and in between for DWA.
The average total Social Force on the ego agent (right-hand plot) across all time steps is highest for DWA, and significantly lower for SF and ORCA agents. 

With respect to the quality/social metrics using PGP significantly improves both the average Space Violation Rate and the average total Social Force (right-hand side) across all (C)CA algorithms. This means, agents with PGP move into the personal space of other pedestrians significantly less often than without PGP. At the same time, they create less "tension", i.e., Social Force when moving through the denser groups of people.

In sum, using PGP to bias the (C)CA algorithms' goals significantly improves the quality of the behavior. At the same time, their performance increases in terms of collision avoidance, and, in the case of DWA, also in terms of time-to-target. This comes at the expense of slightly longer paths, and, for SF agents, at the cost of a marginally longer time-to-target.

\section{PGP on WaPOCHI mobile robot}

\begin{figure*}[h!]
  \centering
  \begin{subfigure}{0.45\textwidth}
    \centering
    \includegraphics[width=\linewidth]{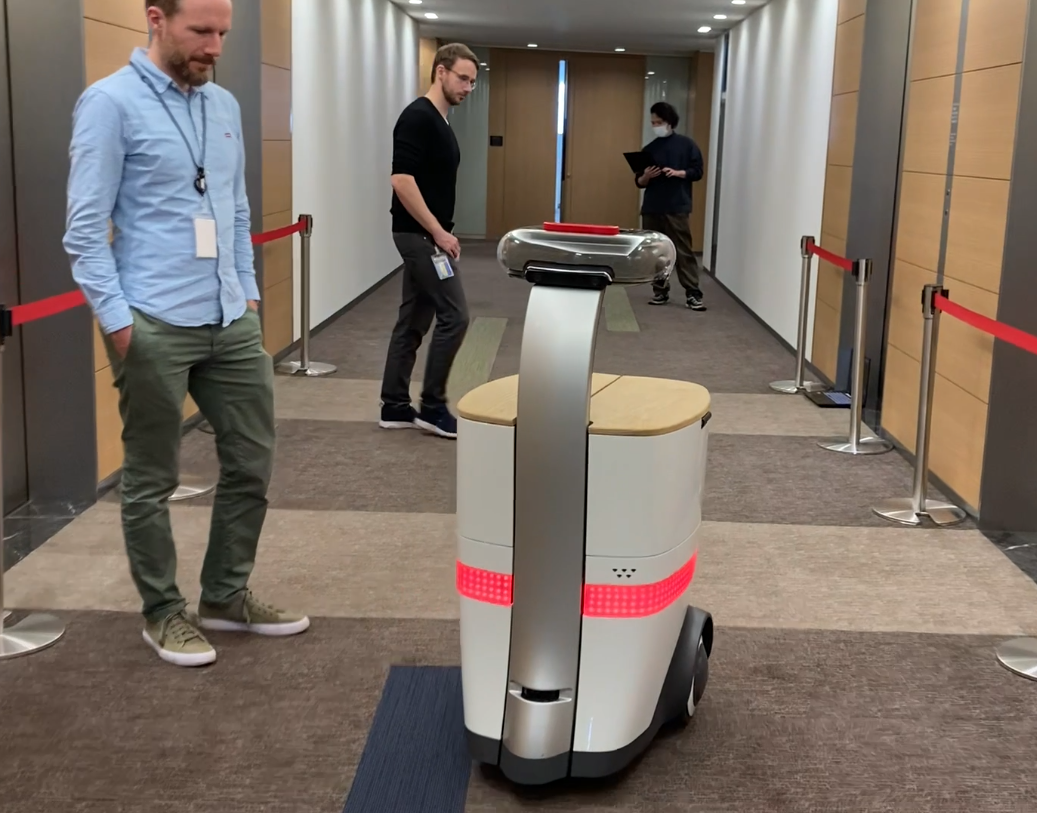}
    \label{pgp_wapochi_image}
  \end{subfigure}
  \hspace {12mm}
  \begin{subfigure}{0.20\textwidth}
    \centering
    \includegraphics[width=\linewidth]{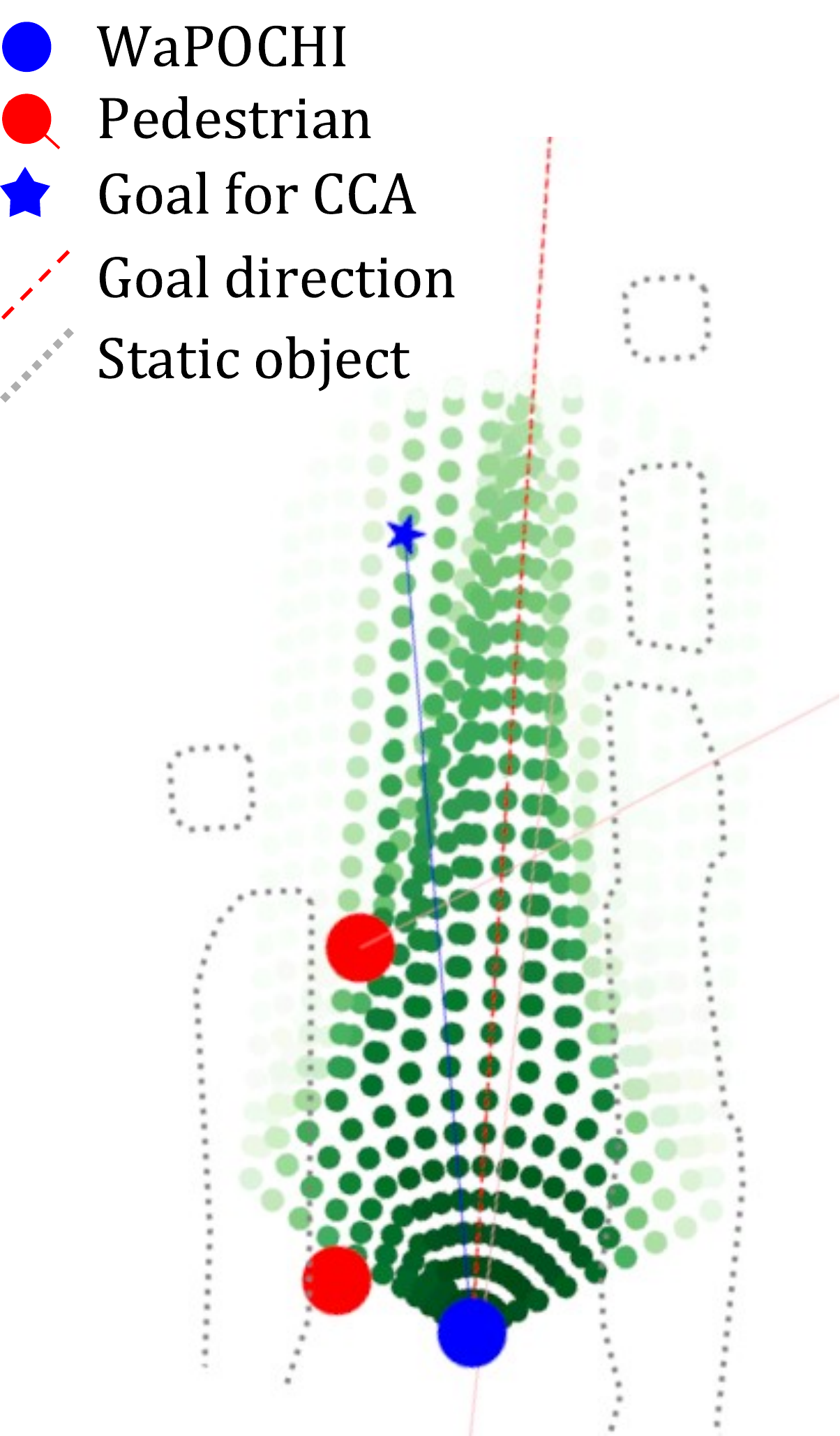}
    \label{pgp_wapochi_mci}
  \end{subfigure}
  
  \caption{Left-hand side: WaPOCHI mobile robot at Honda R\&D, Tokyo, in a corridor scenario with PGP planner. The robot's map goal is at the end of the corridor. Right-hand side: Visualization of sensed input to PGP (grey: static objects, red: pedestrians), PGP's planning (green dots representing different evaluated trajectories, light color indicating low survival probability), and PGP output (blue star and dotted line indicating short-term goal direction to DWA). Note how PGP biases the (C)CA planner towards the left side of the corridor (blue star in right plot), since the person with the black shirt is predicted to move to the right side.}
  \label{pgp_wapochi}
\end{figure*}
\noindent We integrated PGP on the WaPOCHI robot of Honda R\&D, an advanced prototype mobile robot. WaPOCHI is capable of navigating indoor and outdoor public spaces, leading and following registered users as well as carrying items.\footnote{\url{https://global.honda/en/stories/119-2402-hondaci.html}} The WaPOCHI platform offers a 360-degree field of view with stereo and monocular cameras, and it is equipped with an IMU, LiDAR bumpers, powerful computational hardware, and a mobile base. WaPOCHI perception in terms of static and dynamic objects is vision-based. It uses state-of-the-art machine learning models to detect dynamic and static objects.
WaPOCHI's navigation stack contains interchangeable modules, connected by the system's ROS2 framework.

PGP is integrated in WaPOCHI as a ROS2 module, running in 10 Hz. It receives the robot's position and goal, its velocity and orientation, as well as positions and velocities for pedestrians and the boundaries of static objects. In each iteration, it outputs a subgoal to a DWA module for collision avoidance, which plans the final trajectory for the robot passed to the controller. We tested PGP in an indoor corridor scenario (see Figure \ref{pgp_wapochi}). The robot's task was to navigate the corridors with a behavior that is not only collision-free, but foresighted and cooperative. In the tests, PGP improved the robots' behavior by biasing DWA's target towards areas with fewer predicted conflicts. In the example in Figure \ref{pgp_wapochi}, it outputs a goal on the left-hand side of the corridor (blue star in right-hand plot) to pass behind the crossing pedestrian.
This is consistent with the findings of the simulation study, where we see small but statistically significant improvements in space compliance rates for agents with PGP starting already at low densities.
The proof-of-concept integration into WaPOCHI shows that PGP can improve a robot's behavior at relatively low additional computational cost, works with real-world data, and is conceptually simple to add to a mobile robotics platform. 

\section{Conclusion}
\noindent In this paper, we presented the Probabilistic Gap Planner (PGP), a novel trajectory-based long-term planner for Social Robot Navigation. PGP is based on the assumption that interacting pedestrians are, in general, cooperative. The approach decouples conflict avoidance (adjusting the motion behavior early on to traverse less dense areas) from cooperative collision avoidance (finding a collision-free, cooperative motion for an \textit{imminent} interaction). PGP modifies an established probabilistic collision risk model to accommodate the assumption of cooperativity. It creates a set of prototypical trajectory variations leading to the global navigation goal. To evaluate them, PGP uses the so-called \textit{survival analysis} of the risk model to weigh the expected future rewards with the probability of reaching them. We run extensive simulations with over 6.600 runs. We show that PGP guides the agents with state-of-the-art (C)CA algorithms to less crowded areas. Using PGP significantly improves the established quality/social metrics \textit{Space Compliance} and average total \textit{Social~Force}. At the same time, using PGP improves the agent's performance in terms of their collision rate, and, in the case of DWA, also the time-to-target. Typically, these gains come at the expense of a slightly longer total path length, and the additional computational effort to run PGP.
We integrated PGP on the Honda R\&D WaPOCHI mobile robot, running in real-time at 10 Hz in a ROS2 environment. First tests with the robot show that PGP is able to guide the robot towards areas with less pedestrians in a corridor scenario. A promising area of research is improving PGP's trajectory space exploration strategy. Since the simple set of trajectory variations used in this paper already leads to significant improvements in agent behavior, using explorative sampling techniques or a search-based approach could enable the agents to find even better solutions.

\bibliographystyle{IEEEtran}
\bibliography{IEEEabrv,SocialRobotNavigation}
\end{document}